\documentclass{article} 
\usepackage{iclr2021_conference,times}

\usepackage{amsmath,amsfonts,bm}

\def\eqref#1{equation~\ref{#1}}

\def\1{\bm{1}}

\DeclareMathAlphabet{\mathsfit}{\encodingdefault}{\sfdefault}{m}{sl}
\SetMathAlphabet{\mathsfit}{bold}{\encodingdefault}{\sfdefault}{bx}{n}

\usepackage{multirow}
\usepackage{booktabs}
\usepackage{hyperref}
\usepackage{subfigure}
\usepackage{graphicx}
\usepackage[noend]{algpseudocode}
\usepackage{algorithmicx,algorithm}
\usepackage{enumitem}
\usepackage{bbding}
\usepackage{url}

\title{PV-RCNN++: Semantical Point-Voxel Feature Interaction for 3D Object Detection}

\author{Peng Wu\textsuperscript{1}, Lipeng Gu\textsuperscript{1}, Xuefeng Yan*\textsuperscript{1}, Haoran Xie\textsuperscript{2}, \\\textbf{Fu Lee Wang\textsuperscript{3}, Gary Cheng\textsuperscript{4}, Mingqiang Wei\textsuperscript{5}}\\
\textsuperscript{1}Nanjing University of Aeronautics and Astronautics, Nanjing, China\\
\textsuperscript{2}Lingnan University, HongKong, China\\
\textsuperscript{3}Hong Kong Metropolitan University, HongKong, China\\
\textsuperscript{4}The Education University of Hong Kong, HongKong, China\\
\textsuperscript{5}Shenzhen Research Institute, Nanjing University of Aeronautics and Astronautics, Shenzhen, China\\
}

\iclrfinalcopy
\begin{document}

\maketitle

\begin{abstract}
Large imbalance often exists between the foreground points (i.e., objects) and the background points in outdoor LiDAR point clouds.
It hinders cutting-edge detectors from focusing on informative areas to produce accurate 3D object detection results.
This paper proposes a novel object detection network by semantical point-voxel feature interaction, dubbed PV-RCNN++. 
Unlike most of existing methods, PV-RCNN++ explores the semantic information to enhance the quality of object detection. 
First, a semantic segmentation module is proposed to retain more discriminative foreground keypoints.
Such a module will guide our PV-RCNN++ to integrate more object-related point-wise and voxel-wise features in the pivotal areas. 
Then, to make points and voxels interact efficiently, we utilize voxel query based on Manhattan distance to quickly sample voxel-wise features around keypoints. Such the voxel query will reduce the time complexity from \emph{O(N)} to \emph{O(K)}, compared to the ball query. Further, to avoid being stuck in learning only local features, an attention-based residual PointNet module is designed to expand the receptive field to adaptively aggregate the neighboring voxel-wise features into keypoints. 
Extensive experiments on the KITTI dataset show that PV-RCNN++ achieves 81.60$\%$, 40.18$\%$, 68.21$\%$ 3D mAP on Car, Pedestrian, and Cyclist, achieving comparable or even better performance to the state-of-the-arts.
\end{abstract}

\section{Introduction}

Object detection in both 2D and 3D fields \cite{ren2016faster,zhengchengyu22,zongqiwei22,yin2021center,luo2021exploring,ji2022stereo,WangXWLW22}  is increasingly important with the development of autonomous driving \cite{geiger2012we}, robot systems, and virtual reality. Much progress has been made in 3D object detection via various data representation (e.g., monocular images \cite{yan2017mono3d,reading2021categorical,chen2020monopair,li2019gs3d}, stereo cameras \cite{chen20173d,li2019stereo}, and LiDAR point clouds). Compared to 3D object detection from 2D images , LiDAR point cloud casts a critical role in detecting 3D objects as it contains relatively precise depth and 3D spatial structure information.

LiDAR-based 3D object detectors can be roughly grouped into two prevailing categories: Voxel-based \cite{zhou2018voxelnet,yan2018second,ye2020hvnet,lang2019pointpillars,shi2019part} and Point-based \cite{shi2019pointrcnn,shi2020point,yang20203dssd,xie2021venet}. The former discretizes points into regular grids for the convenience of the 3D Sparse Convolutional Neural Network (CNN). Then, the voxelized feature map can be compressed to Bird's Eye View (BEV) which is fed to Region Proposal Network (RPN) \cite{ren2016faster, yan2018second} to produce predictions. On the contrary, the point-based ones mainly adopt PointNet++ \cite{qi2017pointnet++} as the backbone, which take raw points as input, and abstract sets of point features through an iterative sampling-and-grouping operation.
Different from only the single voxel-based and point-based methods, PV-RCNN \cite{shi2020pv} explores the interaction between point-wise and voxel-wise features. 
Specially, PV-RCNN deeply integrates both 3D Voxel CNN and PointNet-based Set Abstraction (SA) to enhance the ability of feature learning. To be concrete, a Voxel Set Abstraction (VSA) is proposed to encode voxel-wise features of different scales through sampled keypoints by Furthest Point Sampling \cite{qi2017pointnet++} (FPS). Through coordinate transform and projection, VSA also concatenates the features of BEV and raw point features into keypoints to have a more comprehensive understanding of 3D scenes.

Nevertheless, we observe that the large imbalance between small informative areas containing 3D objects and large redundant background areas exists in outdoor LiDAR point clouds. It poses a challenge for accurate 3D object detection. 
Generally, the point cloud obtained by LiDAR covers a long range of hundreds of meters, where only several small cars are captured and the rest are numerous background points.
However, in PV-RCNN \cite{shi2020pv}, the whole 3D scene is summarized through a small number of sampled keypoints by FPS. When selecting keypoints, FPS tends to choose distant points to evenly cover the whole point cloud, which causes excessive unimportant background points retained and many valuable foreground points discarded. Consequently, the performance of PV-RCNN is limited largely due to insufficient features provided by the foreground objects. 
Therefore, we consider that if there is any prior knowledge that can lead the detector to focusing on the pivotal foreground objects to extract more valuable features. 
The inspiration is to leverage the result of point cloud semantic segmentation as the prior knowledge to guide the detector.

To this end, we present a novel 3D object detection network via semantical point-voxel feature interaction (termed PV-RCNN++). 
First, we introduce a lightweight and fast foreground point sampling head meticulously modified from PointNet++ \cite{qi2017pointnet++} to select proper object-related keypoints. We remove the feature propagation (FP) layer in PointNet++  to avoid the heavy memory usage and time consumption \cite{yang20203dssd}. We only remain the SA layers to produce more valuable keypoints. Concretely, in each SA layer, we adopt a binary segmentation module to classify the foreground and background points. Then, inspired by \cite{chen2022sasa}, we adopt a novel sampling strategy, semantic-guided furthest point sampling (S-FPS), taking segmentation scores as guide to sample and group representative points. Different from FPS, S-FPS gives more preference to positive points, making more foreground points retained in the SA layers. Hence, the sampled points in the SA layers can act as the pivotal point-wise representation for the succeeding operation.

After obtaining the discriminative keypoints, the challenge would be how to efficiently integrate the voxel-wise and point-wise features via keypoints. We seek to 1) to speed up the interaction between points and voxels; and 2) to effectively summarize the 3D information from voxel-wise feature. Specifically, 3D sparse convolution is first adopted to encode the voxelized point cloud. Then, we propose a fast voxel-to-point interaction module to efficiently sample and group neighboring voxel-wise feature around keypoints. The existing query strategy, ball query \cite{shi2020pv}, consumes too much time to compute the Euclidean distance from every voxel to keypoints to identify whether the voxel is within a given radius or not. Therefore, motivated by \cite{deng2021voxel}, we regard keypoints as voxels, which are regularly arranged in 3D space, and then a voxel query strategy based on Manhattan distance is utilized to quickly identify the neighboring voxel-wise feature of each keypoint. Compared with ball query, our voxel query greatly reduces the time consumption from $O(N)$ to $O(K)$, where $N$ is the total number of voxels and $K$ means the number of neighboring voxels around keypoints.

An Attention-based \cite{vaswani2017attention} Residual PointNet Module is proposed to abstract the neighboring voxel-wise features to summarize the multi-scale 3D information. We apply self-attention mechanism on voxel set of each keypoint to produce corresponding attention maps, allowing each voxel to have a more comprehensive perception field containing more 3D structure and scene information of other nearby voxels. Last, we introduce a lightweight residual \cite{he2016deep,ma2022rethinking} PointNet module to further extract and aggregate the refined voxel-wise feature.

The main contributions are summarized as follows:
\begin{itemize}
	\item We introduce a semantic-guided keypoint sampling module to retain more valuable foreground points from the point cloud, which helps the detector focus on small pivotal areas containing 3D objects.
	\item We utilize voxel query based on Manhattan distance to quickly gather the neighboring voxel-wise features around keypoints, reducing the time consumption compared to ball query and improving the efficiency of point-voxel interaction.
	\item We propose an attention-based residual PointNet module, which allows each voxel to have an adaptive and nonlocal summary of the neighborhood to achieve more accurate predictions.
	\item Extensive experiments and results show that our proposed method achieves comparable performance on common 3D object detection benchmark, KITTI dataset \cite{geiger2012we}.
\end{itemize}

\section{Related Work}\label{sec2}

\textbf{Point-Based Method.}
Generally, point-based methods \cite{shi2019pointrcnn,shi2020point,zarzar2019pointrgcn,yang20203dssd,zhang2021pc} mainly rely on PointNet++ \cite{qi2017pointnet++} and Graph Neural Network \cite{scarselli2008graph}. To preserve the position information, point-based methods operate directly on the point cloud to extract point-wise feature. Inspired by Faster R-CNN \cite{ren2016faster}, PointRCNN \cite{shi2019pointrcnn} utilizes PointNet++ to design a bottom-up 3D Region Proposal Network to generate 3D proposal boxes which are then refined in canonical coordinate at the second stage. To reduce the time-consuming operation of the Feature Propagation(FP) layer in PointNet++, 3DSSD \cite{yang20203dssd} removes the FP layer and proposes a feature-guided sampling method to lead the SA layer to select more discriminative points. Different from common point-based methods, PC-RGNN \cite{zhang2021pc} proposes a Graph-based point cloud completion \cite{yuan2018pcn} module to capture the geometry clues of 3D proposals, which provide more complete structure and shape information for refinement. Although point-based methods have the potential to achieve more accurate detection, the problem of time complexity and memory consumption cannot be settled properly, which limits their further development.

\textbf{Voxel-Based Method.} 
Voxel-based methods \cite{zhou2018voxelnet,yan2018second,ye2020hvnet,lang2019pointpillars,shi2019part,zheng2021cia,zheng2021se} adopt the regular voxelized point cloud as data representation. VoxelNet \cite{zhou2018voxelnet} first proposes a generic single-stage network for 3D box regression. In VoxelNet, Voxel Feature Encoder(VFE) is designed to divide unordered points into grids and utilizes simplified PointNet to produce a representative feature of each voxel. 3D CNN is then applied to extract the feature of the whole 3D scene. CIA-SSD \cite{zheng2021cia} proposes a Spatial-Semantic feature aggregation module to extract low-level spatial features and high-level semantic features of the BEV feature map for more accurate proposals. SE-SSD \cite{zheng2021se} utilizes Knowledge Distillation \cite{hinton2015distilling} to design a pair of teacher and student SSDs. An effective IoU-based matching strategy is proposed to filter soft targets from the teacher and formulate a consistency loss to align student predictions with them. Despite that voxel-based methods improve the speed and efficiency of detection, they pay a price for degrading the localization accuracy due to the loss of point-wise feature.

\textbf{Point-Voxel Hybrid Method.} 
To achieve both efficiency and accuracy, Point-Voxel hybrid methods \cite{yang2019std,he2020structure,shi2020pv} try to take advantage of the efficient computation of voxel-based backbone and accurate position information from raw point clouds. STD \cite{yang2019std} designs a Spherical Anchor to generate proposals from raw point clouds with the help of PointNet-like networks. Then at the second stage, a local VFE is utilized to extract the voxel-wise feature to help box prediction. At the basis of SECOND \cite{yan2018second}, SA-SSD \cite{he2020structure} adds a detachable auxiliary network to transform voxel-wise tensors into the point-wise feature and then leverages point-wise structure information to help train the backbone network. PV-RCNN \cite{shi2020pv} designs a Voxel Set Abstraction module to integrate the multi-scale voxel-wise and point-feature through sampled keypoints. VIC-Net \cite{jiang2021vic} presents a two-branch network, which consists of a point branch for geometry detail extraction and a voxel branch for efficient proposals generation. In this paper, we propose a semantic-guided point-voxel interaction method, PV-RCNN++, to efficiently integrate the voxel-wise and point-wise feature, providing a more comprehensive and discriminative feature for accurate prediction.

\textbf{LiDAR-Camera Fusion Method.}
Recently, many multi-modal works \cite{chen2017multi,ku2018joint,qi2018frustum,huang2020epnet,vora2020pointpainting,yoo20203d,pang2022fast,Zhang_2022_CVPR,wang2022multi} have been proposed to explore the fusion of different sensor data for 3D object detection. MV3D \cite{chen2017multi} is a pioneering work to directly combine the feature from point cloud BEV map, front view map, and 2D images to locate objects. EPNet \cite{huang2020epnet} adopts a refined way in which each point in the point cloud is fused with the corresponding image pixel to obtain more accurate detection. However, all these methods inevitably consume a lot of computation time and memory. Moreover, the alignment error between points and pixels will have a negative impact on the accuracy.

\textbf{Attention Mechanism.}
In the past several years, attention mechanism \cite{vaswani2017attention} has shown its great power in many 2D visual tasks \cite{dosovitskiy2020image,shu2022expansion,shu2021spatiotemporal,tang2022coherence,li2022research} as it can effectively capture nonlocal information for feature extraction. Nowadays, some works \cite{zhao2021point,guo2021pct} apply attention mechanism on 3D point cloud processing (e.g., point cloud segmentation) to obtain outperforming results. However, these methods only explore the attentive relation among points while ignoring the potential effects on voxels. Therefore, we introduce an attention-based voxel aggregation module to adaptively extract the spatial information in the 3D scenes for more accurate 3D object detection. 

\section{Methodology}\label{sec3}
\subsection{Overview}
\begin{figure*}[!ht] \centering
	\includegraphics[width=1\linewidth]{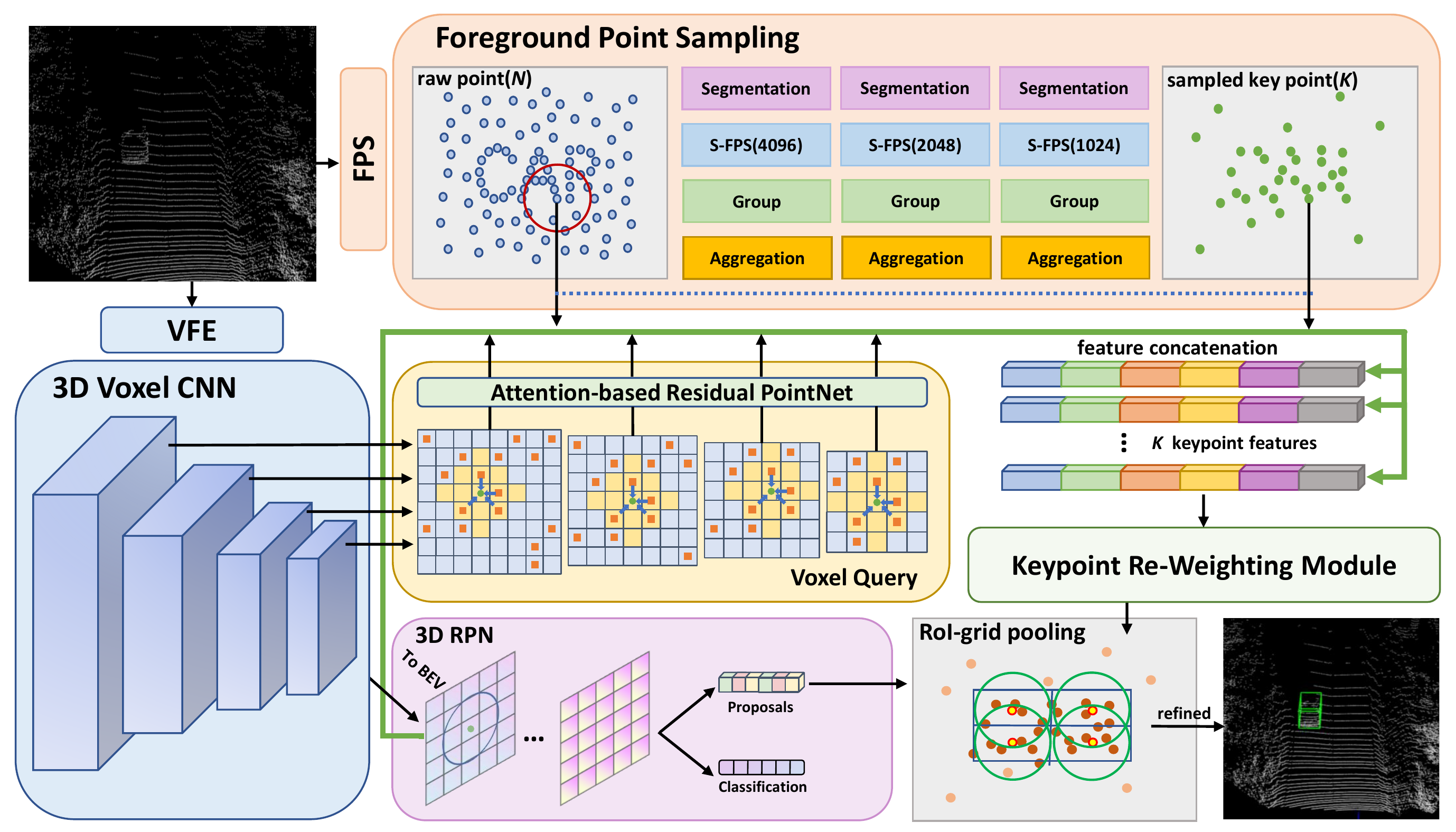}
	\caption{An overview of our PV-RCNN++. The point cloud is first fed into 3D Voxel Encoder to produce 3D proposals. Then, the Foreground Point Sampling module selects more valuable object-related keypoints through the modified SA layers. Last, according to the sampled keypoints, the voxel-wise feature, point-wise feature, and the BEV feature are concatenated to be fed into RoI-grid pooling to refine the proposals to produce more accurate 3D boxes. }
	\label{fig:network framework}
\end{figure*}

Beyond previous wisdom, we argue that 1) more foreground points are beneficial to capturing pivotal structure and position feature; 2) faster query strategy is needed to relieve the time-consuming interaction between points and voxels; and 3) voxel-wise feature should have a more comprehensive perception of neighboring structure feature instead of local convolution feature. To this end, we present our PV-RCNN++: Semantical Point-Voxel Feature Interaction for 3D object detection, which consists of the following modules: 1) a binary segmentation module is introduced to guide FPS to select more object-related keypoints; 2) voxel query based on Manhattan distance replaces the ball query to quickly sample voxel-wise feature; 3) an attention-based residual PointNet is designed to adaptively fuse the neighboring voxel-wise feature to summarize the nonlocal 3D structure information. Our backbone is illustrated as Fig.~\ref{fig:network framework}. Given point cloud $P=\{p_i\mid i=1,2,3...N\} \subseteq $ $\mathbb{R}^{3+d}$ as input, where $N=16384$ and $d$ denotes the point feature (e.g., reflection intensity), our goal is to predict the center location $(x,y,z)$, box size $(l,w,h)$ and rotation angle $\theta$ around $Z-$axis of each object.

\subsection{Voxel Encoder and 3D Region Proposal Network}
 First, the unordered point cloud are transformed to uniformed 3D grids with voxel size $v$, and each grid contains $k$ points. Then, the mean Voxel Feature Encoder(VFE) \cite{zhou2018voxelnet} is adopted to compute the mean feature of $k$ points as the representative feature of the grid.

\textbf{3D Voxel CNN.} Through mean VFE, the point cloud shapes as a  $L \times W \times H$ feature volume. We utilize 3D Sparse Convolutional Neural Networks \cite{yan2018second} to encode the feature volumes with $1\times,2\times,4\times,8\times$ downsample sizes. All four downsampled voxel-wise feature maps are preserved for the subsequent point-voxel interaction module. 

\textbf{3D Region Proposal Network.} After 3D Voxel Encoder, the forth 8$\times$ downsampled feature map is compressed to 2D BEV feature map which is of $\frac{L}{8} \times \frac{W}{8} $ resolution. We utilize anchor-based methods \cite{ren2016faster} on the BEV feature map to generate 3D anchor boxes with the average size of each class. Considering the rotation angle around $Z$-axis, $0$ and $\frac{\pi }{2} $ degree are set for each anchor pixel. Therefore, the whole BEV map produces $3\times 2 \times \frac{L}{8}\times\frac{W}{8}$ proposals in total for three classes: Car, Pedestrian, and Cyclist.
\subsection{Foreground Point Sampling}
\begin{figure*}[!ht] \centering
	\includegraphics[width=1\linewidth]{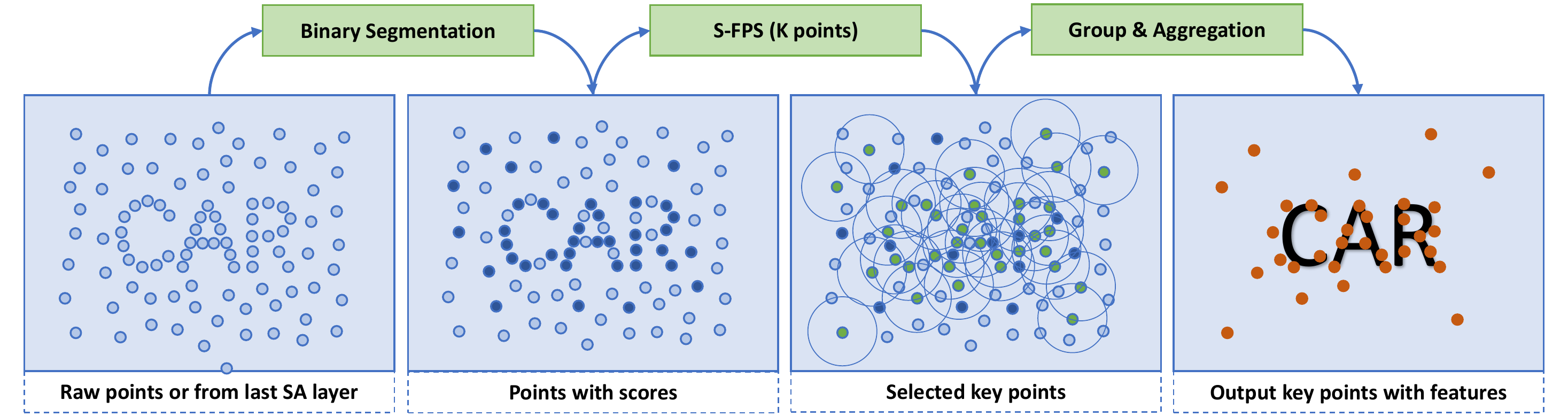}
	\caption{The structure of the modified Set Abstract(SA) layer in the Foreground Point Sampling module. Points from raw data or the previous SA layer are fed to the binary segmentation module to obtain the scores. Then, S-FPS is adopted to select more foreground keypoints to have a better understanding of the 3D structure information. Last, neighboring points around keypoints are grouped and aggregated to produce the final keypoint feature,  which is fed to the next SA layer.  }
	\label{fig:Segmentation}
\end{figure*}

Our motivation is to retain more foreground points to capture more valuable spatial and position information while bringing no burden of time consumption, so we meticulously redesign the PointNet++ to act as our foreground point sampling module. As mentioned in 3D-SSD \cite{yang20203dssd}, despite the FP layer in PointNet++ \cite{qi2017pointnet++} can broadcast the semantic feature to all points to improve segmentation precision, it takes much time to upsample points. Therefore, capturing more foreground points in the SA layer would be a better choice. We remove the FP layer in PointNet++ and adopt a semantic-guided sampling strategy in SA layer. We first feed the raw points $P$ with feature $F$ into the segmentation module to compute the score $S$. Then with the guide of $S$, we employ the modified sampling strategy to sample $K$ keypoints in the SA layer. The specific process is shown in Fig.~\ref{fig:Segmentation} and described as follows.

\textbf{Binary Segmentation Module.} To avoid bringing high computation, we adopt a 2-layer MLP as binary segmentation module to directly obtain the score of point. Concretely, given point feature set $F_k=\{f^{d_k}_1,f^{d_k}_2,f^{d_k}_3,...f^{d_k}_{N_k}\}$, where $d_k$ denotes the $d$-dimension of point feature $f^{d_k}_i$ fed into the $k$-th SA layer, score $s_i\in [0,1]$ of each point is defined as:
\begin{equation}
    s_i = \Omega (\mathcal{MLP}_k(f^{d_k}_i))
\end{equation}
where $\Omega$ means the sigmoid function, $\mathcal{MLP}_k$ represents the segmentation module in $k$-th SA layer. The real segmentation labels can be obtained from ground-truth boxes. We define the points inside the ground-truth boxes as foreground points while the outside as background points. Therefore the loss of the segmentation module can be calculated as:
\begin{equation}
    \mathcal{L}_{seg}^k =\sum_{k=1}^{m} \frac{\lambda _k}{N_k} \cdot \sum_{i=1}^{N_k} BCE(s_i^k,\hat s_i^k)
\end{equation}
where $s_i^k$ denotes the predicted score and $\hat s_i^k$ is the ground-truth label (0 for the background and 1 for the foreground) of the $i$-th point in the $k$-th SA layer. $BCE$ is the binary cross-entropy loss function. $m$ means the number of SA layers, which is set to 4. $N_k \in \{4096,2048,1024,256\}$ is the total number of the input points and $\lambda _k\in\{0.1,0.01,0.001,0.0001\}$ is the loss weight of $k$-th SA layer.

\textbf{Semantic-guided Further Point Sampling.}
Since we have obtained the point scores from the binary segmentation module, it means that the possible foreground points have been masked. The easiest way to select foreground points is to use Top-K scores as a guide. However, as observed from Fig.~\ref{fig:visual sampling strategy} and Table~\ref{tab2}, it will decrease the perceptual ability of the whole 3D scene if too many foreground points are selected and few background points are involved. Motivated by \cite{chen2022sasa}, we modified the furthest point sampling strategy by adding a score weight called S-FPS. Keeping the basic flow of FPS unchanged, we leverage scores of unselected points to rectify the distances to the selected points. Given point coordinate set $\mathbf{P}=\{p_1,p_2,p_3,...p_{N}\}\subseteq $ $\mathbb{R}^{3}$ and the corresponding score set $\mathbf{S}=\{ s_1,s_2,s_3,...s_{N}\}$, the distance set $\mathbf{D}=\{d_1,d_2,d_3,...d_N\}$ is the shortest distances of $N$ unselected points to the already selected points. In the original FPS, the point with the longest distance is picked up as the furthest point. However, in S-FPS, we recalculate each distance $\Hat d_i$ with score $s_i$ as the following formula:
\begin{equation}\label{gamma}
    \Hat d_i = (e^{\gamma s_i}-1)\cdot d_i,
\end{equation}
where $\gamma$ is an adjustable parameter deciding the importance of the score, which is set to 1 by default. When $\gamma$ is fixed, the closer score $s_i$ is to 1, the greater $\Hat d_i$ is. Hence, S-FPS can select more positive points (i.e., foreground points) compared to FPS. Obviously, S-FPS will be similar to Top-K algorithm if $\gamma=+\infty $. The specific process of S-FPS is shown in Algorithm 1.

\begin{algorithm}\label{SFPS}
\caption{Semantic-guided Furthest Point Sampling. $N$ is the number of input points. $M$ is the number of sampled keypoints.} 
{\hspace*{0.02in} \bf Data:} 
Coordinates of Points $\mathbf{P}=\{p_1,p_2,...p_N\}$;\\
\hspace*{0.46in} 
Scores of Points $\mathbf{S}=\{ s_1,s_2,...s_{N}\}$\\
{\hspace*{0.02in} \bf Result:} 
Sampled keypoints $\mathbf{K}=\{ k_1,k_2,...k_{M}\}$
\begin{algorithmic}[1]
\State initialize an empty sampled points set $\mathbf{K}$
\State initialize a distance array $\mathbf{D}=\{d_1,d_2,...d_N\}$ with all $+\infty$ 
\State initialize a flag array $\mathbf{F}=\{f_1,f_2,...f_N$\} with all zeros
\For{$i=1$ to $M$}
    \If{$i=1$} 
        \State $k_i$ = argmax($\mathbf{S}$)
    \Else
        \State $\mathbf{D}=\{(e^{\gamma s_k}-1)\cdot d_k\mid f_k=0\}$
        \State $k_i$ = argmax($\mathbf{D}$)
    \EndIf
    \State add $k_i$ to $\mathbf{K}$, $f_{k_i}=1$
    \For{$j=1$ to $N$}
        \State $d_j$ = min$(d_j,\parallel p_j - p_{k_i}\parallel)$   
    \EndFor
\EndFor
\State \Return $\mathbf{K}$
\end{algorithmic}
\end{algorithm}

\subsection{Faster Neighboring Voxel Group}
The original neighboring voxel sampling strategy in PV-RCNN \cite{shi2020pv} employs ball query \cite{qi2017pointnet++} to sample 3D voxel feature around keypoints. However, the ball query occupies too much time to compute the Euclidean distance between each voxel and keypoint, which is rather low efficiency. It is worth noting that the voxels are regularly arranged in ordered 3D space, which can be easily accessed by indices. Motivated by Voxel R-CNN \cite{deng2021voxel}, we apply the Manhattan distance to replace the Euclidean distance to compute the query distance. Differently, we directly operate on keypoints while Voxel R-CNN operates on 3D proposal grids, which makes our approach take less computation but focus more on critical information. 
\begin{figure*}[!ht] \centering
	\includegraphics[width=0.5\linewidth]{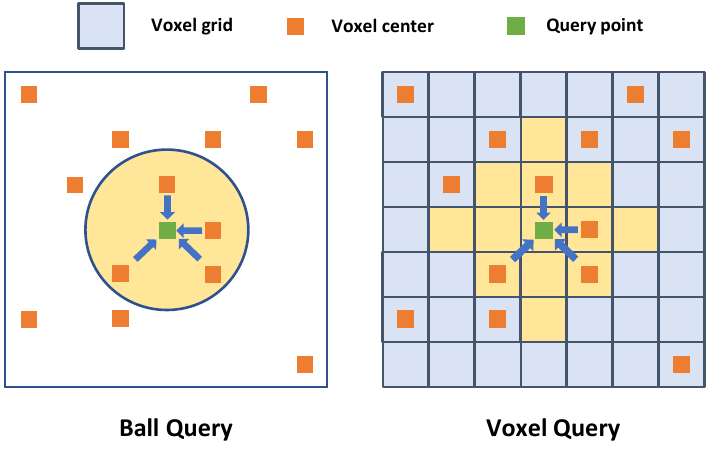}
	\caption{Demonstration of Ball query and Voxel query in 2D view (performed in 3D space).}
	\label{fig:voxel query}
\end{figure*}

\textbf{Point to Voxel Coordinate. }
The coordinates of selected keypoints must be transformed to voxel coordinates to query voxel indices in the corresponding 3D voxel feature map. Given point coordinate $[x_p,y_p,z_p]$, voxel size $[v_x,v_y,v_z]$, point cloud range $[x_{min},x_{max},y_{min},y_{max},z_{min},z_{max}]$, and 3D CNN downsample stride $c_k$, the voxel coordinate $[x_v,y_v,z_v]$ in the $k$-th 3D voxel feature map can be calculated as follows:
\begin{equation}
    [x_v,y_v,z_v]= [\frac{x_p-x_{min}}{v_x\cdot c_k }, \frac{y_p-y_{min}}{v_y\cdot c_k }, \frac{z_p-z_{min}}{v_z\cdot c_k }, ]
\end{equation}

\textbf{Voxel Query.}
Compared to ball query, voxel query utilize a positive integer $I$ (default $I=4$) as query range to produce offsets $\{\bigtriangleup_x,\bigtriangleup _y,\bigtriangleup _z\mid \bigtriangleup\in [-I,+I]\}$ to access the neighboring voxels around center voxel $[x_v,y_v,z_v]$. To determine whether a neighboring voxel is within the query radius, a distance radius threshold is set to compute the Manhattan distance. To be concrete, given the center voxel coordinate $[x_v,y_v,z_v]$ and queried voxel coordinate $[x_q,y_q,z_q]$, the Manhattan distance $d_{man}$ can be computed as:
\begin{equation}
    d_{man} = \mid x_v - x_q \mid + \mid y_v - y_q \mid + \mid z_v - z_q \mid
\end{equation}
To sample $M$ nearby voxels in total $N$ voxels, the ball query needs to compute distance $N$ times while the voxel query only needs $K$ times $(M<K<N)$, where $K$ is the number of neighboring voxels around the center voxel, reducing the time complexity from $O(N)$ to $O(K)$.

\subsection{Attention-based Residual PointNet}
In previous work \cite{shi2020pv}, a simple PointNet-like MLP is directly adopted to aggregate the coarse feature of sampled voxels. Nevertheless, each queried voxel feature contributes unequally to learning the local structure information. 3D sparse convolution operation limits the voxel's ability to better understand the neighboring 3D structure and semantic information. Therefore, we consider how to adaptively focus on the crucial voxel-wise feature to obtain more comprehensive and important features.

Attention mechanism \cite{vaswani2017attention} has shown its great power in various visual tasks. Benefiting from the self-attention mechanism, the model can obtain a larger receptive field to summarize the nonlocal feature. We note that sampled voxel-wise feature by voxel query can provide different structure and spatial information from different areas of the point cloud. Hence, we propose an Attention-based Residual PointNet Aggregation module to adaptively aggregate the voxel-wise feature from the hotspot area of the point cloud.

\begin{figure*}[!ht] \centering
	\includegraphics[width=0.65\linewidth]{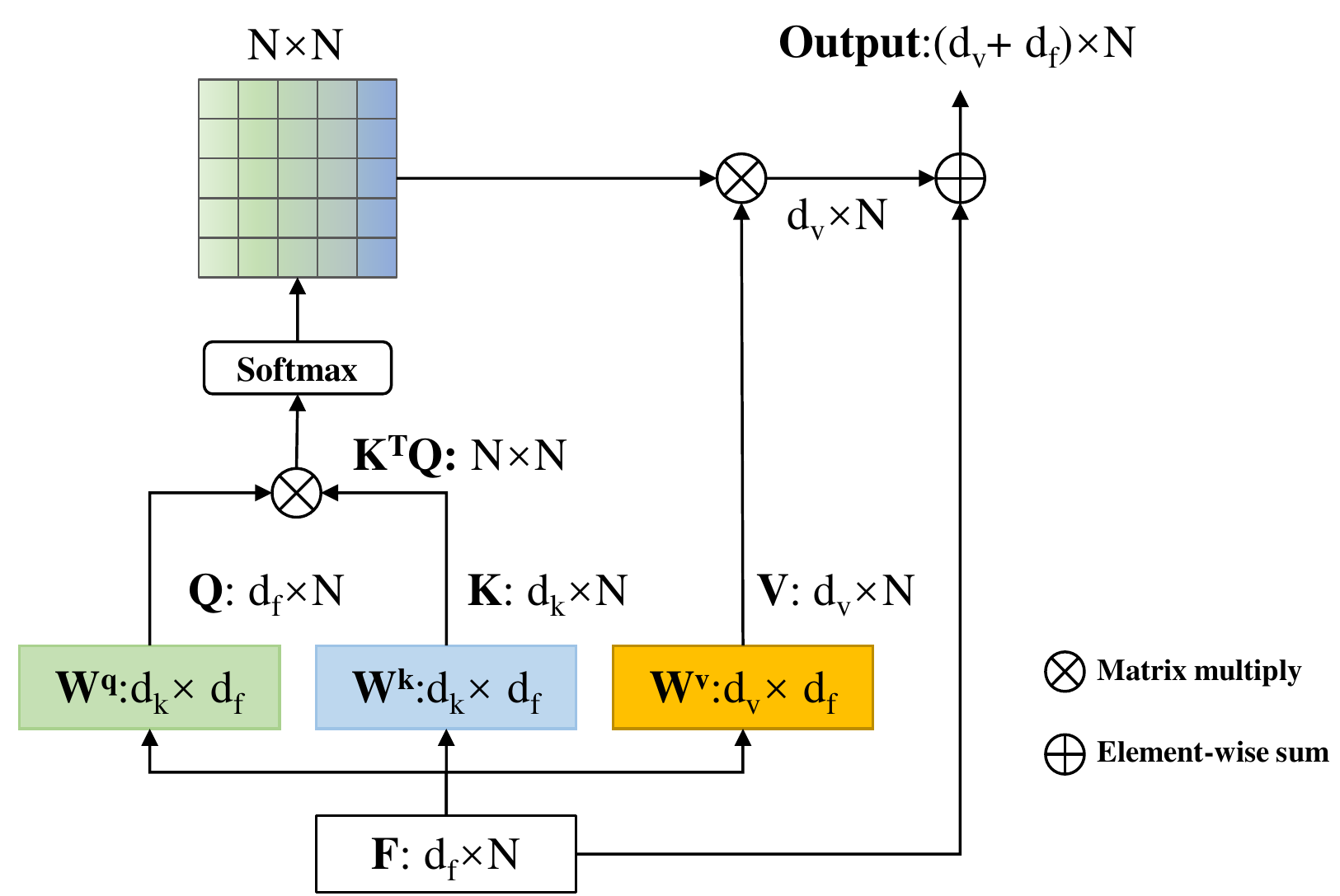}
	\caption{Demonstration of Voxel Attention Module. The input is $N$ sampled voxel-wise feature.}
	\label{fig:Attention}
\end{figure*}

\textbf{Voxel Attention Module.}
As shown in Fig.~\ref{fig:Attention}, given feature set $\mathbf F = \{ f_1,f_2,f_3,...f_N\}\in \mathbb{R}^{d_f\cdot N }$ of $N$ sampled voxels by voxel query, queries $\mathbf{Q}$, keys $\mathbf{K}$ and values $\mathbf{V}$ are generated from $\mathbf{F}$ as following formula:
\begin{equation}
    \mathbf{Q}=\mathbf{W}^q\mathbf{F}\in\mathbb{R}^{d_k\cdot N},\
    \mathbf{K}=\mathbf{W}^k\mathbf{F}\in\mathbb{R}^{d_k\cdot N},\
    \mathbf{V}=\mathbf{W}^v\mathbf{F}\in\mathbb{R}^{d_v\cdot N},
\end{equation}
where $ \mathbf{W}^q \in\mathbb{R}^{d_k\cdot d_f},  \mathbf{W}^k \in\mathbb{R}^{d_k\cdot d_f}, \mathbf{W}^v \in\mathbb{R}^{d_v\cdot d_f} $ are linear projections consisting of learnable matrices. Then the attention weights $\mathbf{S}_i=\{s^i_1,s^i_2,...,s^i_N \mid i\in[1,N]\}$ of the $i$-th query are calculated by softmax function on dot-product similarity between keys $\mathbf{K}_i$ and queries $\mathbf{Q}_i$:

\begin{equation}
    \mathbf{S}_i = softmax(\frac{ \mathbf{K}^T_i\mathbf{Q}_i }{ \sqrt{d_k} } ),
\end{equation}
where $d_k$ is a scaling factor, which is set to the length of the voxel-wise feature dimension. Since we have got the attention weights of each query, then we generate the fine-grained values $\mathbf{\hat{V}}_i \in\mathbb{R}^{d_v}$ by computing the weighted sum of $\mathbf{V}\in\mathbb{R}^{d_v\cdot N}$:
\begin{equation}
    \mathbf{\hat{V}}_i = Attention(\mathbf{Q}_i,\mathbf{K},\mathbf{V})=\sum_{m=1}^{N}\mathbf{S}_i^m\cdot \mathbf{V}^m.
\end{equation}
Finally, we add the weighted value $\mathbf{\hat{V}}\in\mathbb{R}^{d_v\cdot N}$ and original voxel-wise feature  $\mathbf{F}\in\mathbb{R}^{d_f\cdot N}$ to represent the attention feature $\mathbf{\tilde{V}}\in\mathbb{R}^{(d_v+d_f)\cdot N}$.

\begin{figure*}[!ht] \centering
	\includegraphics[width=0.6\linewidth]{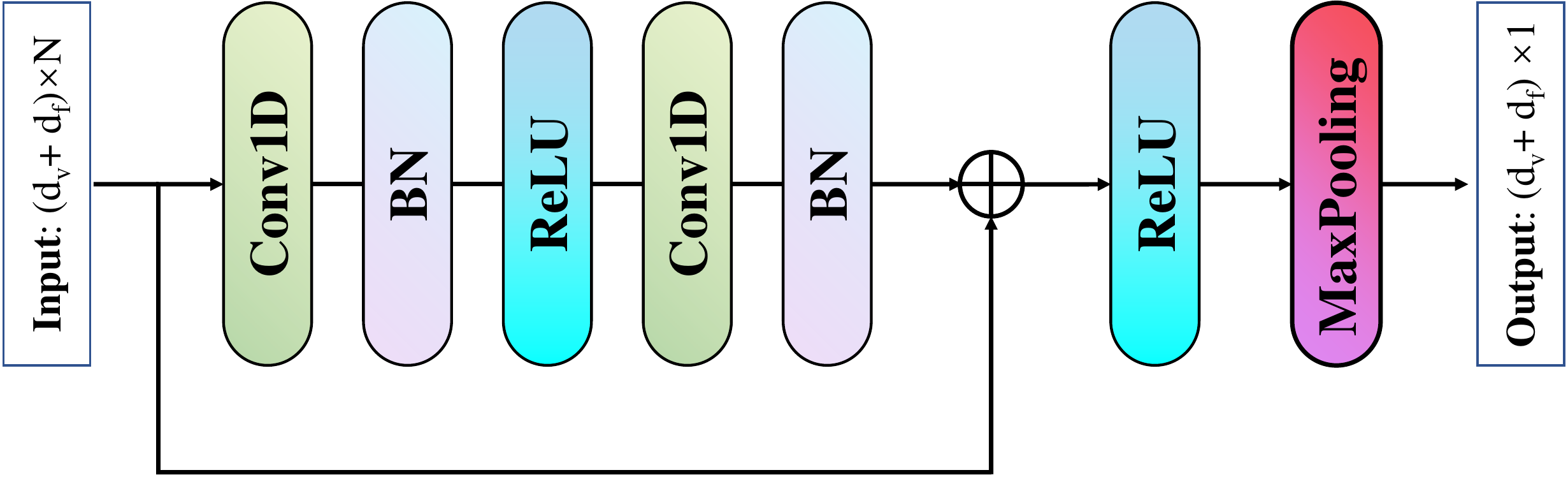}
	\caption{Demonstration of Residual PointNet Aggregation. It consists of simple Convolution1D, Batch Normalization, Relu, and MaxPooling.}
	\label{fig:residual}
\end{figure*}

\textbf{Residual PointNet Aggregation.}
Through the self-attention mechanism, each voxel point integrates the weighted feature from surrounding voxels and can adaptively focus on the hotspot area of local structure information. Next, the weighted values $\mathbf{\tilde{V}}= \{v_1,v_2,v_3,...v_N\}$ are fed into a feed-forward network to produce the final feature. Different from Transformer \cite{vaswani2017attention}, which adopts simple linear layers, we propose a lightweight residual PointNet to forward the weighted values. As illustrated in Fig.~\ref{fig:residual}, given weighted values $\mathbf{\tilde{V}}$, a plug-and-play module composed of Convolution1D, Batch Normalization and Relu is stacked in a skip-connection way to extract the feature of $\mathbf{\tilde{V}}$. In the end, we adopt the MaxPooling function to produce the representative voxel aggregation feature.

\subsection{RoI Grid Pooling}
As shown in Fig.~\ref{fig:network framework}, besides summarizing the voxel-wise feature, keypoints $\mathbf{P}=\{p_1,p_2,p_3,...p_k\}$ also concatenate the grouped feature from raw points and BEV features, which make keypoints rich in 3D spatial feature and semantic feature. Next, all the keypoint feature $\hat {\mathbf{F}}=\{f_1,f_2,f_3,...f_N\}$ are fed to the second stage, RoI-grid pooling, to refine the 3D proposals generated by RPN to achieve more accurate and robust results. RoI-grid pooling uniformly divides each 3D proposal into $6\times6\times6$ grids which can be denoted as $\mathbf{G} =\{g_1,g_2,g_3,...g_{216}\}$. Benefiting from our foreground sampling module, the keypoint feature contains more object-related feature, which makes each grid $g_i$ contain more informative foreground keypoints. Specifically, given a radius $\widetilde{r}$ and grid point $g_j$, the feature $f_i$ of keypoint $p_i$ is grouped if the keypoint is within $\widetilde{r}$. The grouped keypoint feature set $\mathbf{K}$ is defined as follows:
\begin{equation}
    \mathbf{K} = \{ [f_i;p_i-g_j]^T \mid \forall p_i\in \mathbf{P},\forall g_j\in \mathbf{G}, \parallel p_i-g_j\parallel \ < \widetilde{r},\}
\end{equation}
where $p_i-g_j$ is the relative location from $p_i$ to $g_j$, which is concatenated to the feature $f_i$. Then the grouped keypoint feature set $\mathbf{K}$ is fed to a PointNet-like \cite{qi2017pointnet++} module to produce a refined feature. After obtaining the refined RoI-grid feature of each proposal, a 2-layer MLP is adopted to vectorize the RoI-grid feature to 256 dimensions to represent the final feature.
\subsection{Loss Function}
Our method is an end-to-end trainable network, which is optimized by a multi-task loss $\mathcal{L}_{total}$ as follows:
\begin{equation}
    \mathcal{L}_{total} = \mathcal{L}_{seg} + \mathcal{L}_{rpn}+\mathcal{L}_{rcnn} + \mathcal{L}_{key}
\end{equation}
As we have mentioned in Sec 3.3, the segmentation loss $\mathcal{L}_{seg}$ is computed by binary cross-entropy loss on sampled keypoints in SA layers. Similar to \cite{yan2018second}, $\mathcal{L}_{rpn}$ is composed of three partial loss:
\begin{equation}
    \mathcal{L}_{rpn} = \alpha_1 \mathcal{L}_{cls}+\alpha_2 \mathcal{L}_{loc}+\alpha_3 \mathcal{L}_{dir}
\end{equation}
where $\alpha_1, \alpha_2, \alpha_3 $ are assigned $\{1.0, 2.0, 0.2\}$ to represent the weight coefficient of object classification loss $\mathcal{L}_{cls}$, location regression loss $\mathcal{L}_{loc}$, and direction regression loss $\mathcal{L}_{dir}$ respectively. To avoid the case in which our model is stuck in determining the directions of objects, we give  $\alpha_3$ a relatively small parameter. To be concrete, $\mathcal{L}_{cls}$ is computed by focal loss introduced by RetinaNet \cite{lin2017focal} due to the large imbalance between the foreground and background classes. $\mathcal{L}_{loc}$ is optimized by smooth-L1 loss function for box regression and $\mathcal{L}_{dir}$ is computed by sine-error loss \cite{yan2018second} for angle regression.

$\mathcal{L}_{rcnn}$ \cite{shi2020pv} is the loss of between RoIs and ground-truth labels, which consists of the classification confidence loss $\mathcal{L}_{rcnn\_cls}$, location regression loss $\mathcal{L}_{rcnn\_loc}$, and box corner loss $\mathcal{L}_{rcnn\_corner}$ in the refinement stage. The computation of $\mathcal{L}_{rcnn}$ is defined as follows:
\begin{equation}
    \mathcal{L}_{rcnn} = \mathcal{L}_{rcnn\_cls}+ \mathcal{L}_{rcnn\_loc}+\mathcal{L}_{rcnn\_corner}
\end{equation}
where $\mathcal{L}_{rcnn\_cls}$ is optimized with the binary cross-entropy loss function, and $\mathcal{L}_{rcnn\_loc}$ and  $\mathcal{L}_{rcnn\_corner}$ are optimaized smooth-L1 loss function as in \cite{shi2020pv,yan2018second}.

\begin{figure*}[!ht] \centering
	\includegraphics[width=0.4\linewidth]{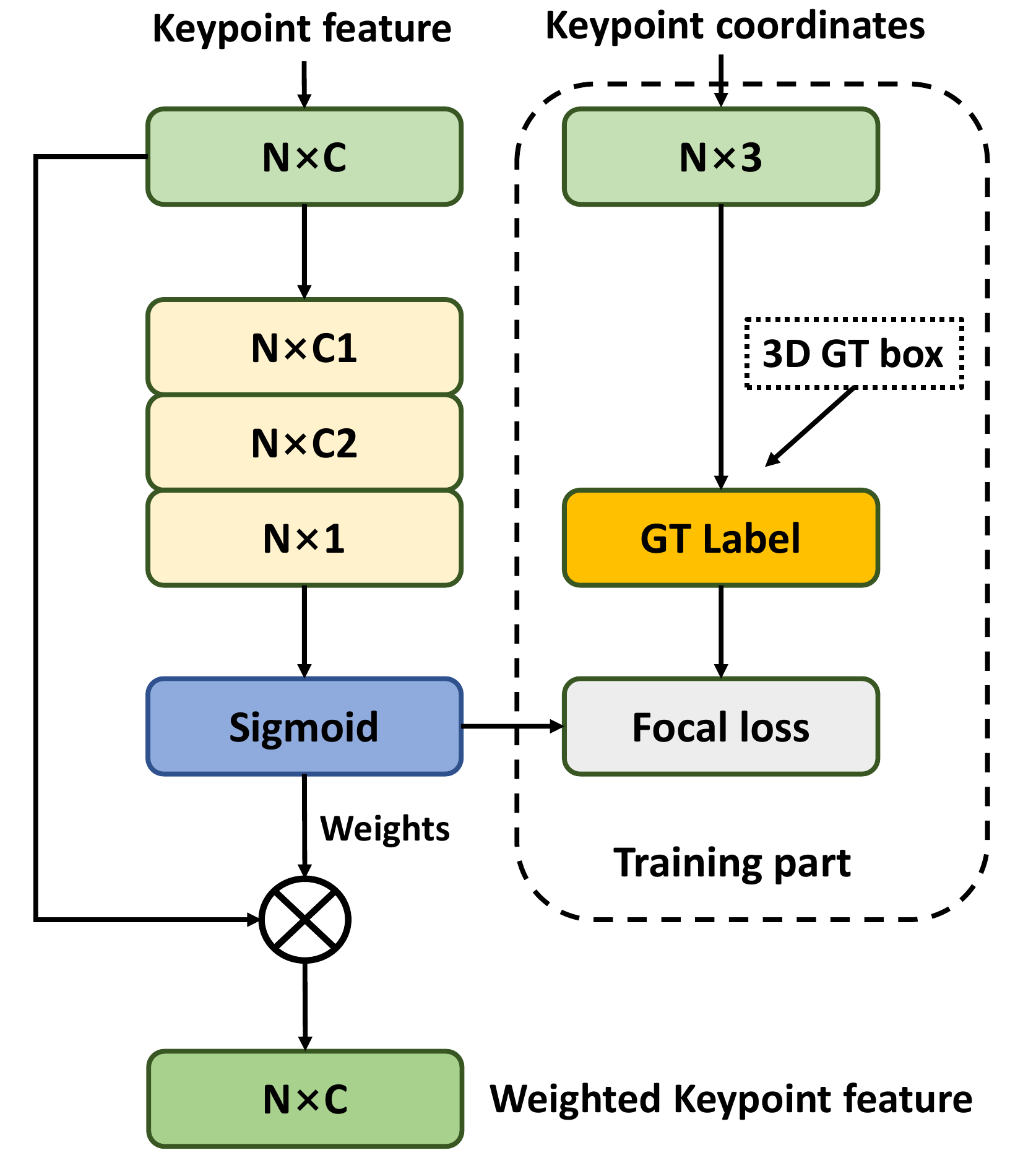}
	\caption{Demonstration of Keypoint Re-weighting Module. }
	\label{fig:KRW}
\end{figure*}

Besides, after the whole 3D scene is summarized into a small number of keypoints, it is reasonable to re-weight them to ensure that the foreground point feature has greater weights to contribute more to the refinement stage. As demonstrated in Fig.~\ref{fig:KRW}, the re-weighting loss of keypoints $\mathcal{L}_{key}$ is computed by the focal loss \cite{lin2017focal} between predicted keypoint scores and true labels in ground-truth boxes. 
\section{Experiments}
\subsection{Datasets}
\textbf{KITTI.} 
KITTI Dataset \cite{geiger2012we} is one of the most popular benchmarks for 3D object detection in autonomous driving. It collects 7481 LiDAR point cloud frames for training and 7518 for testing. Although LiDAR scans the $360^{\circ}$ point cloud scene, only the objects in Field of Vision (FOV) are annotated with 3D boxes. Specifically, the training dataset is divided into train split (3712 samples) and val split (3769 samples) for our experiments. When submitting the test results to the KITTI server, we train 90$\%$ training data to obtain a robust and highly generalized model.

\subsection{Implementation Details}
\textbf{Voxelization.} 
For KITTI Dataset, we only use the FOV point cloud as raw data. The range of FOV field is $[0,70.4]$ meters along $X$-axis, $[-40,40]$ meters along $Y$-axis and $[-3,1]$ meters along $Z$-axis, which is regularly divided into regularly arranged voxels with voxel size $0.05m\times0.05m\times0.1m$.

\textbf{Sampling Strategy Setting.} 
In the foreground point sampling module, we place four SA layers for summarizing the semantic feature with different numbers of keypoints. Before SA layers, we sample 16384 points from the raw points as input. For the reason that raw points have no segmentation scores, we adopt the original FPS in the first SA layer to sample 4096 keypoints. Then, in the next three SA layers, we utilize S-FPS to sample 2048, 1024, and 256 keypoints according to the segmentation scores.

\textbf{Network Architecture.}
As shown in Fig.~\ref{fig:network framework}, the point cloud is input in the form of quantitative voxels with the resolution of $1600 \times 1408 \times 40 \times 8$ and sampled 16384 points by FPS. First, the voxel-based backbone leverages 3D Voxel CNN with $1\times,2\times,4\times,8\times$ downsample sizes to produce four voxel-wise feature maps with 16, 32, 64, and 64 output dimensions. Following the last 3D CNN layer, the output feature map is of the resolution of $200 \times 176 \times 2 \times 128$. Once the feature map is reshaped to $200 \times 176 \times 256$, the RPN \cite{yan2018second} network is applied to generate 3D proposals. For the Voxel Query in Sec 3.4, we set the query range $I$ to 4 and sample 16 neighboring voxels around query point. Hence, the size of each attention map in Voxel Atttention Module is $16\times16$. In the residual PointNet, the output dimension of Convolution1D is set to 32. The 16384 raw points are input to the Foreground Point Sampling module which consists of 4 subsequent SA layers. Each SA layer has 2 multi-scale radii $r\in\{ [0.1m,0.5m], [0.5m,1.0m], [1.0m,2.0m], [2.0m,4.0m]\}$ to group neighbouring points which are input to 2-layer MLP to classify the foreground and background. Then, the sampled keypoints from SA layer integrate the point-wise feature, voxel-wise feature, and 2D BEV feature to summarize the critical information of the 3D scene. In the end, RoI-grid pooling divides the proposal box into $6\times6\times6$ grids to refine the proposals with the aggregation of feature-rich keypoints within 0.6m and 0.8m radius.

\textbf{Training and Inference.}
Concretely, the model is trained 80 epochs with Adam optimizer and the learning rate is initially set to 0.01 updated by one-cycle policy. We conduct the experiments on one RTX 3090 GPU with batch size 4, which takes about 36 hours.

During training, to avoid over-fitting, we leverage data augmentation strategies like \cite{yan2018second}. We use GT-sampling to paste some foreground instances from other point cloud scenes to current training frame. Besides, augmentation operations like random flip along $X$-axis, random rotation with angle range in $[-\frac{\pi}{4},\frac{\pi}{4}]$, and random scaling with a random scaling factor in $[0.95,1.05]$ are adopted to enhance the generalization and robustness of the model. At the RoI refinement stage, we sample 128 proposals from RPN and set the threshold $\theta_{fg}=$ 0.55 to classify the foreground objects and background objects. Half of them are recognized as foreground objects if the 3D Intersection over Union(IoU) with the ground-truth box is over $\theta_{fg}$ while the lower is determined to be the background.

At the inference stage, we employ non-maximum-suppression(NMS) on the 3D proposals with the threshold of 0.7 to filter out top-100 proposals as the input of the refinement module. Then, in RoI Grid Pooling, proposals are further refined with the feature-rich keypoints. Subsequently, NMS is adopted again with a threshold of 0.1 to remove redundant predictions.
\subsection{Results on KITTI}
\textbf{Evaluation Metrics.}
We follow the evaluation criteria that the KITTI benchmark provides to ensure accuracy and fairness. The IoU threshold is set to 0.7 for Car and 0.5 for both Pedestrian and Cyclist. The results reported to the official KITTI test server are calculated by average precision(AP) setting of recall 40 positions to compare with the state-of-the-art methods.

We demonstrate the test results returned from the KITTI test server in Table \ref{tab1} with the comparison of performance on 3D detection. For the most critical Car detection, we surpass the PV-RCNN by $2.55\%, 0.20\%, 0.07\%$ on easy, moderate, and hard levels. It is worth noting that our method improves greatly on the Cyclist class, surpassing PV-RCNN by $4.89\%, 5.09\%,4.36\%$ on three levels. However, our method achieves inferior results on the Pedestrian class compared with the state-of-the-art methods, where we think the segmentation module has limited ability to classify the small foreground objects like Pedestrian, and S-FPS tends to select keypoints on the big foreground objects like the Car and Cyclist.

\begin{table}[ht]
\centering
\begin{minipage}{\columnwidth}
\caption{Performance comparison on KITTI official test server. The 3D Average Precision is calculated by 40 recall positions.  }\label{tab1}
\resizebox{\columnwidth}{!}{
\begin{tabular}{c|c|c|ccc|ccc|ccc}
\toprule
\multirow{2}{*}{Method} & \multirow{2}{*}{Reference} & \multirow{2}{*}{Modality} & \multicolumn{3}{c|}{Car 3D Detection(\%)} & \multicolumn{3}{c|}{Ped 3D Detection(\%)} & \multicolumn{3}{c}{Cyclsit 3D Detection(\%)}  \\
             & &     & Easy  & Mod   & Hard    & Easy  & Mod   & Hard   & Easy  & Mod   & Hard  \\ 
\hline
MV3D \cite{chen2017multi}  &CVPR2017  & L+C          & 74.97 & 63.63 & 54.00                        & -     & -     & -      & -     & -     & -       \\
AVOD-FPN \cite{ku2018joint}    &CVPR2018              & L+C          & 83.07 & 71.76 & 65.73              & 50.46 & 42.27 & 39.04            & 63.76 & 50.55 & 44.93          \\
F-PointNet \cite{qi2018frustum}   &CVPR2018   & L+C           & 82.19 & 69.79 & 60.59                     & 50.53 & 42.15 & 38.08                     & 72.27 & 56.12 & 49.01   \\
EPNet \cite{huang2020epnet}  &CVPR2020    & L+C         & 89.81 & 79.28 & 74.49                     & -     & -     & -              & -     & -     & -                               \\
PointPainting \cite{vora2020pointpainting}  &CVPR2020         & L+C      & 82.11 & 71.70 & 67.08                     & 50.32 & 40.97 & 37.87                     & 77.63 & 63.78 & 55.89   \\
3D-CVF \cite{yoo20203d}   &CVPR2020   & L+C          & 89.20 & 80.05 & 73.11          & -     & -     & -            & -     & -     & -              \\ 
FAST-CLOCs \cite{pang2022fast}   &CVPR2022       & L+C           & 89.11 & 80.34 & 76.98           & 52.10 & 42.72 & 39.08           & 82.83 & 65.31 &  57.43       \\
CAT-Det \cite{Zhang_2022_CVPR} &CVPR2022 & L+C         &\textbf{89.87} & 81.32 & 76.68         & \textbf{54.26}& \textbf{45.44} &\textbf{ 41.94 }        & \textbf{83.68} & \textbf{68.81}& \textbf{61.45}   \\
\hline
SECOND-V1.5 \cite{yan2018second}   &Sensors2018   & L        & 84.65 & 75.96 & 68.71   & -     & -     & -        & -     & -     & -       \\
PointPillars \cite{lang2019pointpillars}   &CVPR2019         & L          & 82.58 & 74.31 & 68.99      & 51.45 & 41.92 & 38.89         & 77.1  & 58.65 & 51.92         \\
Part-A$^2$ \cite{shi2019part}    &CVPR2019    & L          & 87.81 & 78.49 & 73.51                     & 53.10 & 43.35 & 40.06                     & 79.17 & 63.52 & 56.93       \\
PointRCNN \cite{shi2019pointrcnn}   &CVPR2019   & L          & 86.96 & 75.64 & 70.7                      & 47.98 & 39.37 & 36.01              & 74.96 & 58.82 & 52.53    \\
STD \cite{yang2019std}     &ICCV2019    & L            & 87.95 & 79.71 & 75.09                     & 53.29 & 42.47 & 38.35                     & 78.69 & 61.59 & 55.3               \\
SA-SSD \cite{he2020structure}    &CVPR2020   & L           & 88.75 & 79.79 & 74.16                     & -     & -     & -               & -     & -     & -         \\
3DSSD \cite{yang20203dssd}   &CVPR2020      & L    & 88.36 & 79.57 & 74.55       & \textbf{54.64} & \textbf{44.27} & 40.23         & 82.48 & 64.1  & 56.90        \\
CIA-SSD \cite{zheng2021cia}    &AAAI2021   & L            & 89.59 & 80.28 & 72.87                     & -     & -     & -                         & -     & -     & -              \\
VIC-Net \cite{jiang2021vic}  &ICRA2021   & L   & 88.25 & 80.61 & 75.83                     & 43.82  & 37.18  & 35.35           & 78.29    & 63.65   &57.27    \\

HVPR \cite{Noh_2021_CVPR}    &CVPR2021      & L         & 86.38 & 77.92 & 73.04                     & 53.47    & 43.96     & \textbf{40.64}         & -     & -     & -     \\
SVGA-Net \cite{he2022svga}                  &AAAI2022    &L                &87.33 & 80.47& 75.91                      & 48.48 &40.39& 37.92                       & 78.58& 62.28 &54.88 \\
IA-SSD \cite{zhang2022not}  &CVPR2022   & L             &88.34 & 80.13 &75.04                      &46.51 &39.03 &35.60                       & 78.35 & 61.94 &55.70             \\
\hline
PV-RCNN \cite{shi2020pv}  &CVPR2020    & L            & 87.98 &\textbf{81.40} & \textbf{77.00}                     & 44.61 & 39.04 & 36.89                     & 78.57 & 63.12 & 56.81        \\
Ours    &-     & L              & \textbf{90.53} & \textbf{81.60} & \textbf{77.07}       & 45.94 & 40.18 & 37.28     & \textbf{83.46} & \textbf{68.21} & \textbf{61.17}    \\
Improvements    &-    & -           & +2.55 & +0.20 & +0.07                     & +1.33 & +1.14 & +0.39                     & +4.89 & +5.09 & +4.36          \\
\bottomrule
\end{tabular}}
\footnotetext{\textbf{Note}: \textbf{L+C} denotes LiDAR-Camera fusion methods, and \textbf{L} represents the LiDAR-only methods. The result of PV-RCNN is obtained from the KITTI server tested with the local model trained on the official source code: OpenPCDet \cite{openpcdet2020}. The top two results are in bold. }

\end{minipage}
\end{table}

\subsection{Ablation Studies}
To comprehensively verify the effectiveness of our method, we conduct ablation studies on Foreground Point Sampling, Voxel query, Voxel Attention Module, and Residual PointNet, respectively. 

\begin{figure*}[!ht] \centering
	\includegraphics[width=1\linewidth]{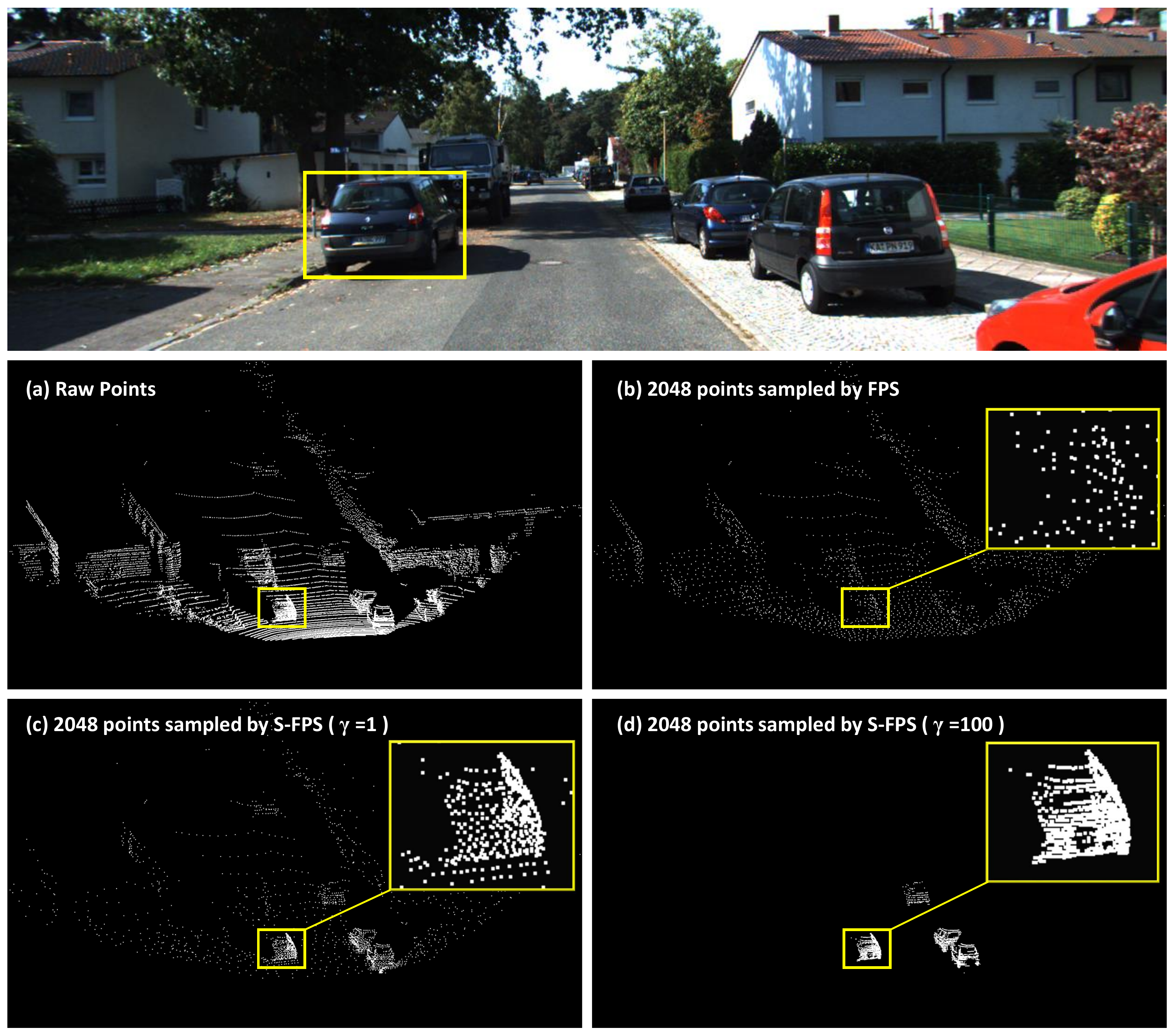}
	\caption{Visualization comparison of different point sampling strategies. (a) shows raw point cloud scene. (b) shows that FPS samples too many background points while points on foreground objects are too sparse. (c) demonstrates that S-FPS$(\gamma=1)$ can sample more foreground points and remains proper background information. (d) shows that almost only foreground points are sampled by S-FPS$(\gamma=100)$, which loses the perception of the whole 3D scene.}
	\label{fig:visual sampling strategy}
\end{figure*}

\textbf{Effectiveness of Foreground Point Sampling. } 
In this part, we test the influence of the different numbers of foreground points on detection precision, as shown in Table~\ref{tab2}. It seems feasible that the more foreground points get caught, the more accurate the result will be. Therefore, we set $\gamma$ in Equation \ref{gamma} with 1, 2, 3, and 100 to increase the number of sampled foreground points. However, it turns out that when $\gamma$ becomes larger, the performance decrease instead. The reason is that sampling excessive foreground points with few background points (like the Top-K sampling strategy) makes it hard for the model to have a global perception of the whole 3D scene, which limits its ability to precisely locate the correct objects. As demonstrated in Fig.~\ref{fig:visual sampling strategy}, when $\gamma=1$, the sampled keypoints can focus on foreground objects and preserve proper background points at the same time.

\begin{table}[ht]
\begin{center}
\caption{Performance comparison on the KITTI val split with different $\gamma$ controlling the number of foreground keypoints. The 3D Average Precision is calculated by 11 recall positions for the Car class. }\label{tab2}
\begin{tabular}{c|c|c|c}
\toprule
Sampling Strategy & Easy(\%)  & Mod(\%) & Hard(\%)\\
\midrule
FPS    & 89.02   & 83.59  & 78.49  \\
S-FPS($\gamma=1$)    & 89.65   & \textbf{84.34}  & \textbf{79.11} \\
S-FPS($\gamma=2$)    & \textbf{89.75}   & 84.15  & 79.09 \\
S-FPS($\gamma=3$)    & 89.69   & 84.11  & 79.02 \\
S-FPS($\gamma=100$)  & 89.06   & 79.06  & 76.74 \\
\bottomrule
\end{tabular}
\end{center}
\end{table}

\textbf{Effectiveness of Voxel Query.}
The function of Voxel Query is to boost the speed of voxel sampling while bringing no loss of performance. Table~\ref{tab3} shows the 3D detection precision for Car class and frame per second (FPS) at the condition of ball query and voxel query. We only replace the ball query with the voxel query while keeping other modules unchanged. Results show that the voxel query improves the inference speed but causes no harm to precision.

\begin{table}[ht]
\begin{center}

\caption{Performance and FPS comparison between ball query and voxel query. The 3D AP is computed on the KITTI val split by recall 11 positions for the Car class.}\label{tab3}%
\begin{tabular}{c|c|c|c|c}
\toprule
Query Method  & Easy(\%)  & Mod(\%) & Hard(\%) & FPS(Hz)\\
\midrule
Ball Query    & 89.42   & \textbf{83.60}  & 82.34 & 12.5 \\
Voxel Query   &\textbf{89.60}   & 83.58  & \textbf{82.45} & \textbf{15.15}\\
\bottomrule
\end{tabular}
\end{center}
\end{table}

\begin{figure*}[!ht] \centering
	\includegraphics[width=1\linewidth]{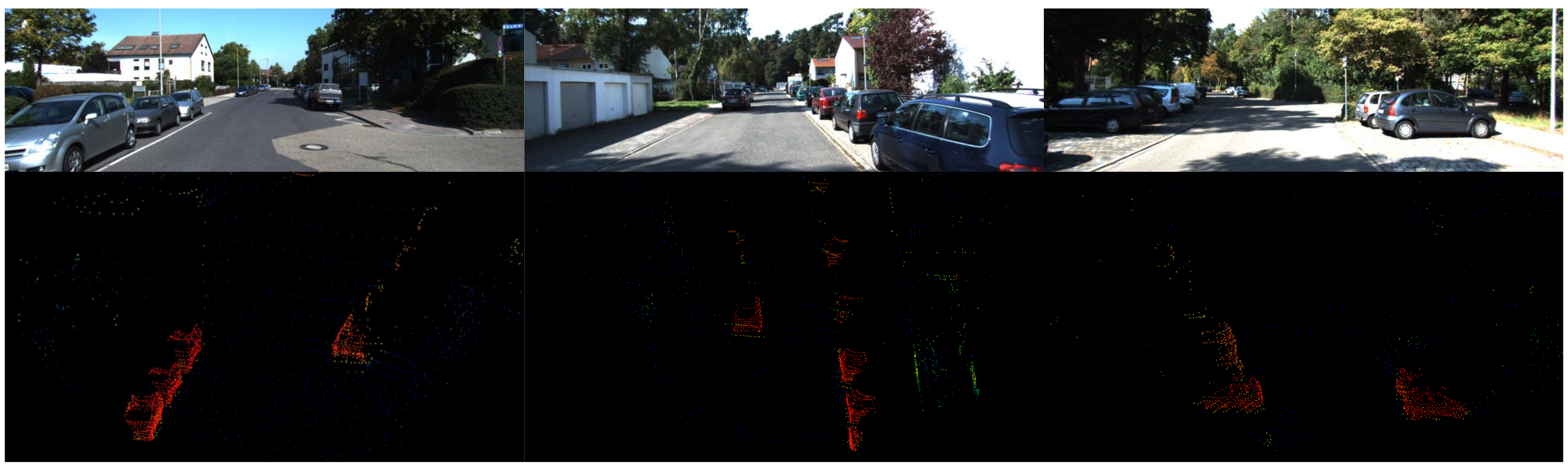}
	\caption{Visualization of attention feature induced by attention-based residual PointNet module. The aggregated features cover the whole object-related regions (red points) rather than small local parts of the object. }
	\label{fig:Attention_maps}
\end{figure*}

\textbf{Effectiveness of Attention-based Residual PointNet.}
We propose the Voxel Attention Module so that voxel-wise feature can obtain a larger receptive field to integrate other voxels' feature instead of the only local feature of itself. Besides, a lightweight residual PointNet module is added to efficiently aggregate the voxel-wise feature. As demonstrated in Table~\ref{tab4}, the performance is further improved by attention-mechanism with residual PointNet. Fig.~\ref{fig:Attention_maps} shows the voxel-wise attention features encoded into keypoints, where almost the whole object is highly focused on, not only local spatial features are learned, indicating the importance of the attention-based residual PointNet module.

\begin{table}[ht]
\centering
\caption{Performance demonstration of Attention-based Residual PointNet Aggregation. The 3D AP is computed on the KITTI val split by recall 11 positions for the Car class.}\label{tab4}%
\resizebox{\textwidth}{!}{
\begin{tabular}{ccccccc}
\toprule
S-FPS & Voxel Query & VAM & Residual PointNet  & Easy(\%)  & Mod(\%) & Hard(\%)\\
\midrule
\XSolidBrush & \XSolidBrush & \XSolidBrush & \XSolidBrush & 89.02  & 83.59 & 78.49  \\
\CheckmarkBold & \CheckmarkBold & \XSolidBrush & \XSolidBrush & 89.60  & 84.08 & 78.98  \\
\CheckmarkBold & \CheckmarkBold & \CheckmarkBold & \XSolidBrush & 89.64  & 84.27 & 79.07 \\
\CheckmarkBold & \CheckmarkBold & \CheckmarkBold & \CheckmarkBold & \textbf{89.65}  & \textbf{84.34} & \textbf{79.11} \\
\bottomrule
\end{tabular}
}
\end{table}

\begin{figure*}[!ht] \centering
	\includegraphics[width=1\linewidth]{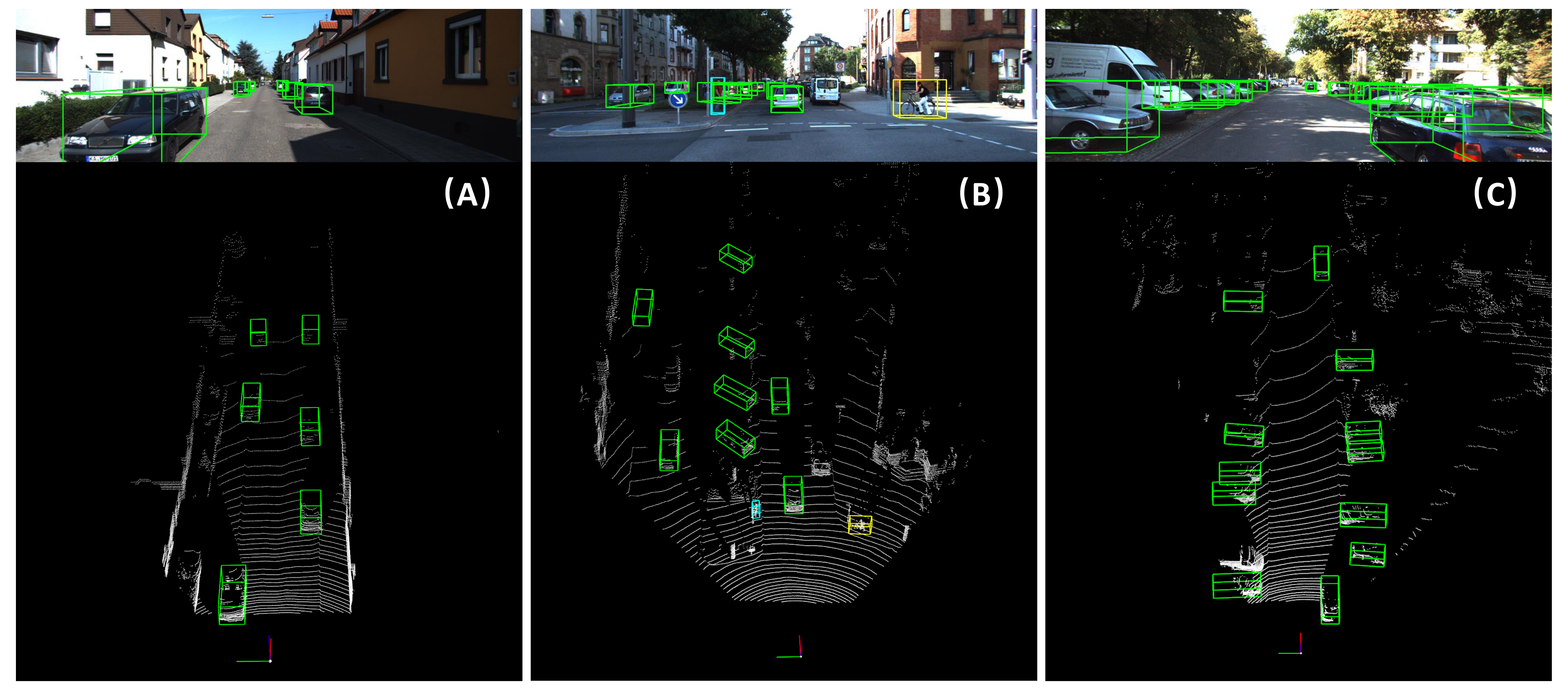}
	\caption{Visualization results on KITTI(FOV). Detected cars, pedestrians, and cyclists are marked by green boxes, blue boxes and yellow boxes.}
	\label{fig:Visualization}
\end{figure*}

\subsection{Visualization and Discussion}
Fig.~\ref{fig:Visualization} shows the visualization results on the KITTI dataset. Our model performs stably and makes accurate detection in complex situations, especially in  Fig.~\ref{fig:Visualization} (B). Moreover, as shown in Table~\ref{tab1}, higher average precision on car and cyclist proves the effectiveness of our proposed methods. However, our method still lags behind some state-of-the-art methods for detecting small objects such as pedestrian. The reason is that our segmentation module performs badly to classify small objects, resulting in fewer keypoints on pedestrians retained to integrate valuable feature for refinement. 
\section{Conclusion}

In this paper, we present the PV-RCNN++ framework, a novel Semantical Point-Voxel Feature Interaction for 3D object detection, which achieves 81.60$\%$, 40.18$\%$, 68.21$\%$ 3D mAP on Car, Pedestrian, and Cyclist on the KITTI benchmark. We introduce a carefully designed point cloud segmentation module as guide to sample more object-related keypoints. Through fast voxel query based on Manhattan distance, we speed up the interaction between keypoints and voxels to efficiently group the neighboring voxel-wise feature. The proposed attention-based residual PointNet abstracts more fine-grained 3D information from nearby voxels, providing more comprehensive features for the succeeding refinement. Extensive experiments on KITTI dataset demonstrate that our proposed semantic-guided voxel-to-keypoint detector precisely summarizes the valuable information from pivotal areas in the point cloud and improves performance compared with previous state-of-the-art methods.

\bibliography{iclr2021_conference}
\bibliographystyle{iclr2021_conference}

\end{document}